\documentclass{article}

\usepackage{PRIMEarxiv}

\usepackage[utf8]{inputenc} 
\usepackage[T1]{fontenc}    
\usepackage{booktabs}       
\usepackage{amsfonts}       
\usepackage{nicefrac}       
\usepackage{microtype}      
\usepackage{lipsum}
\usepackage{fancyhdr}       
\usepackage{graphicx}       
\graphicspath{{media/}}     

\usepackage{url}

\usepackage{graphicx}
\usepackage{subfigure}
\usepackage{wrapfig}
\usepackage{amsthm}
\usepackage{amsmath}
\usepackage{mathtools}
\usepackage{graphicx}
\usepackage{subcaption}
\usepackage{tabularx}
\usepackage{dsfont}
\usepackage{multirow}
\usepackage{wrapfig}
\usepackage{amssymb}
\usepackage{multirow}
\usepackage{color}
\usepackage{booktabs}
\usepackage{gensymb}
\usepackage{enumitem}

\newcolumntype{C}[1]{>{\centering\arraybackslash}p{#1}}

\usepackage{algorithm}
\usepackage{algpseudocode}  

\usepackage[numbers,sort&compress]{natbib}
\usepackage[hidelinks]{hyperref}
\usepackage[capitalize,noabbrev]{cleveref}

\usepackage[table]{xcolor} 

\usepackage{todonotes}

\definecolor{mydarkblue}{rgb}{0,0.08,0.45}
\definecolor{myblue}{HTML}{3b75c9}
\definecolor{myred}{HTML}{E33222}
\definecolor{mygreen}{HTML}{438773}
\definecolor{mymaroon}{RGB}{142,27,19}
\definecolor{maroon}{HTML}{992000}
\definecolor{mycite}{cmyk}{0.55,1,0,0.15}
\definecolor{codeblue}{rgb}{0.25,0.5,0.5}
\definecolor{codekw}{rgb}{0.85, 0.18, 0.50}
\definecolor{codegreen}{rgb}{0,0.6,0}
\definecolor{codegray}{rgb}{0.5,0.5,0.5}
\definecolor{codepurple}{rgb}{0.58,0,0.82}
\definecolor{backcolour}{rgb}{0.95,0.95,0.92}
\definecolor{refcolor}{HTML}{9F363A}
\definecolor{prompt1}{HTML}{3b75c9}
\definecolor{prompt2}{HTML}{9F363A}
\definecolor{prompt3}{rgb}{0.48,0,0.32}
\definecolor{prompt4}{HTML}{438773}
\definecolor{prompt5}{HTML}{6B6B6B}
\hypersetup{
    colorlinks=true,
    citecolor=refcolor,
    linkcolor=myblue
          }

\newcommand{\athena}[1]{\href{https://athena.ohdsi.org/search-terms/start}{Athena Ontology Database}}

\title{Recontextualizing Famous Quotes for Brand Slogan Generation
}

\author{
  Ziao Yang$^{1*}$, Zizhang Chen$^{2*}$, Lei Zhang$^2$ and Hongfu Liu$^1$ \\
  $^1$Brandeis University,  $^2$Adobe\\
  $^*$Equal Contribution\\
  \texttt{\{ziaoyang,hongfuliu\}@brandeis.edu},\{zizhangc,lzhang\}@adobe.com  \\
}

\begin{document}
\maketitle

\begin{abstract}
Slogans are concise and memorable catchphrases that play a crucial role in advertising by conveying brand identity and shaping public perception. However, advertising fatigue reduces the effectiveness of repeated slogans, creating a growing demand for novel, creative, and insightful slogan generation. While recent work leverages large language models (LLMs) for this task, existing approaches often produce stylistically redundant outputs that lack a clear brand persona and appear overtly machine-generated. 
We argue that effective slogans should balance novelty with familiarity and propose a new paradigm that recontextualizes persona-related famous quotes for slogan generation. Well-known quotes naturally align with slogan-length text, employ rich rhetorical devices, and offer depth and insight, making them a powerful resource for creative generation. Technically, we introduce a modular framework that decomposes slogan generation into interpretable subtasks, including quote matching, structural decomposition, vocabulary replacement, and remix generation. Extensive automatic and human evaluations demonstrate marginal improvements in diversity, novelty, emotional impact, and human preference over three state-of-the-art LLM baselines.
\end{abstract}

\section{Introduction}
A slogan—a concise and memorable catchphrase used in advertising—plays a crucial role in communicating the core message or values of a brand, organization, or campaign. It helps capture attention, evoke emotional connections, and distinguish one message from others. A strong slogan reinforces identity, enhances recognition, and makes ideas easier to recall, which is crucial for effectively shaping public perception and sustaining board audience engagement.

Due to advertising fatigue~\citep{abrams2007personalized}, the effectiveness of even a strong slogan diminishes over time as users are repeatedly exposed to it. Consequently, there is a growing demand for novel, creative, and insightful slogans to meet users’ increasingly high expectations. To address this need, designers have begun to leverage generative AI and large language models (LLMs) for slogan creation. For example, \citet{alnajjar2021computational} generates metaphorical slogan candidates based on a target concept and its properties. \citet{ahmad2024enhancing} propose a generative adversarial framework to improve the coherence and diversity of generated slogans. Similarly, \citet{jin2023towards} employs a sequence-to-sequence model for slogan generation with the goal of reducing factual inaccuracies. \citet{hakala2025using}  investigates whether ChatGPT could provide the basis for a constructive slogan invention process for a city.

While prior studies demonstrate the potential of AI-assisted slogan generation, two key limitations remain. First, the generated slogans often fail to clearly reflect the brand persona, resulting in outputs that lack distinctiveness and feel interchangeable across prompts. For example, ChatGPT-generated slogans for Coca-Cola—such as “\textit{Ignite your energy with Coca-Cola’s spark},” “\textit{Boost your vibe with every drink},” and “\textit{Refresh your spirit with a spark}”—exhibit similar phrasing patterns and stylistic redundancy. Second, many generated slogans appear overtly machine-produced, lacking creativity, originality, and insightful impact.

To address these challenges, we revisit the essence of a slogan as a concise and memorable catchphrase and explicitly connect it to its core goals—creativity, originality, and insightfulness. While a completely novel phrase can be difficult to remember, slogans become more effective when they are meaningfully associated with familiar expressions or well-known quotes. Such associations strike a balance between novelty and familiarity, making slogans both creative and easy to recall. Moreover, many famous quotes share a similar length with slogans and naturally employ rich rhetorical devices—such as metaphor, contrast, and special patterns—while also offering depth and insight.

Motivated by these observations, we recontextualize persona-related famous quotes for slogan generation. For example, by remixing Percy Bysshe Shelley’s “\textit{If winter comes, can spring be far behind?}” with Coca-Cola, we obtain the slogan “\textit{Work is coming; Coke cannot be far behind.}” This slogan evokes a widely recognized literary reference, creates a contrast between work and refreshment, frames Coca-Cola as a reward, and ultimately enhances the sense of well-being associated with the brand. As another example, combining Lu Xun’s well-known quote, “\textit{There are two trees in front of my house: one is a jujube tree, and the other is also a jujube tree},” with the Nike brand yields the slogan “\textit{Buy two pairs of shoes: one is Nike, and the other is also Nike.}” This slogan draws on a shared cultural reference familiar to many readers, while effectively conveying strong brand loyalty and reinforcing Nike’s dominant position. Here we summarize our key contributions as follows:
\begin{itemize}[wide=10pt, leftmargin=*, nosep]
    \item Conceptually, we propose a novel paradigm for slogan generation that remixes famous quotes with a brand. Unlike conventional AI approaches, which rely on large corpora of existing slogans for training, our method expands the scope to incorporate well-known quotes, thereby injecting additional creativity and insightfulness into the generated slogans.\vspace{2mm}
    \item Technically, we propose a systematic framework for recontextualizing persona-related famous quotes for slogan generation. Rather than relying on large language models (LLMs) to perform the entire remix process end to end, we deliberately decompose slogan generation into a sequence of well-defined subtasks—quote matching, structural decomposition, vocabulary replacement, and remix generation. Each subtask corresponds to a concrete and interpretable step in the remixing pipeline, enabling fine-grained control over creativity, structure, and semantic alignment. This modular design not only improves the controllability and traceability of the generation process but also helps ensure consistently high-quality and persona-aligned slogan outputs suitable for real-world applications.\vspace{2mm}
    \item We conduct extensive experiments and human evaluation to demonstrate the effectiveness of our proposed recontextualizing framework by remixing famous quotes over three state-of-the-art LLMs in terms of diversity and novelty, emotional impact, and human preference. 
\end{itemize}

\section{Related Work}
Slogan generation~\citep{murakami2023natural}, also known as advertising text generation, can be broadly categorized into template-based and data-driven approaches. Template-based methods~\citep{alnajjar2021computational,bartz2008natural,fujita2010automatic,thomaidou2013automated} rely on predefined linguistic templates to construct slogans. While these approaches ensure grammatical correctness and readability, they often suffer from limited diversity and creativity, resulting in slogans that are repetitive and less appealing to users. In contrast, data-driven methods have become the dominant paradigm in recent years. These approaches leverage machine learning models to learn slogan patterns directly from existing datasets. As a text generation task, slogan generation has been explored using various neural architectures, including LSTM~\citep{hughes2019generating}, RNN~\citep{ahmad2024enhancing}, and Transformer-based models~\citep{liang2023comparative,jin2023towards,kim2023effective}. Although these methods are capable of producing fluent and syntactically coherent slogans, many generated outputs still lack originality, persuasive power, or brand distinctiveness, and occasionally degenerate into generic or low-impact expressions. As a result, they often fail to meet the high standards required for effective advertising content.

To improve the practicality of generated slogans in real-world systems, several studies have explored slogan generation from different perspectives, often in combination with reinforcement learning techniques. \citet{liao2025rlmr} employs mixed reward functions to encourage creativity in generated text, while \citet{wang2025beyond} focuses on enhancing diversity. Other works incorporate user engagement signals: \citet{chen2025ctr} and \citet{wei2022creater} optimize slogan generation with respect to click-through rate, and \citet{kanungo2021ad} and \citet{wang2021reinforcing} aim to improve the attractiveness of generated advertisements. In addition, \citet{golobokov2022deepgen} presents a large-scale system deployed in production for automatically generating sponsored advertisements at web scale.

Recently, there has been a growing surge in employing large language models (LLMs) for slogan generation~\citep{liu2025llms,mita2024striking,meguellati2024good,hakala2025using,jiang2025improving}. While LLMs demonstrate impressive capabilities across many language-generation tasks, the quality of LLM-generated slogans still falls short of practical requirements. A key reason lies in the fundamentally data-driven nature of these approaches: generated slogans are largely learned from existing slogan corpora. As a result, the outputs often resemble prior slogans and suffer from diminished novelty and appeal due to advertising fatigue, and lack engagement.

To address this limitation, we step outside the conventional paradigm of generating slogans solely from existing samples and instead aim to jointly enhance creativity, originality, and insightfulness under a unified framework. Specifically, we leverage famous quotes as a rich source of rhetorical structure and semantic depth, and remix them with brand products to generate slogans. This approach not only avoids expensive model training but also preserves the inherent insightfulness of well-known quotes, while the remixing process introduces additional creativity, surprise, and attractiveness.

\section{Method}
\begin{wrapfigure}[18]{r}{0.4\textwidth} 
  \centering
  \vspace{-15mm}\includegraphics[width=0.38\textwidth]{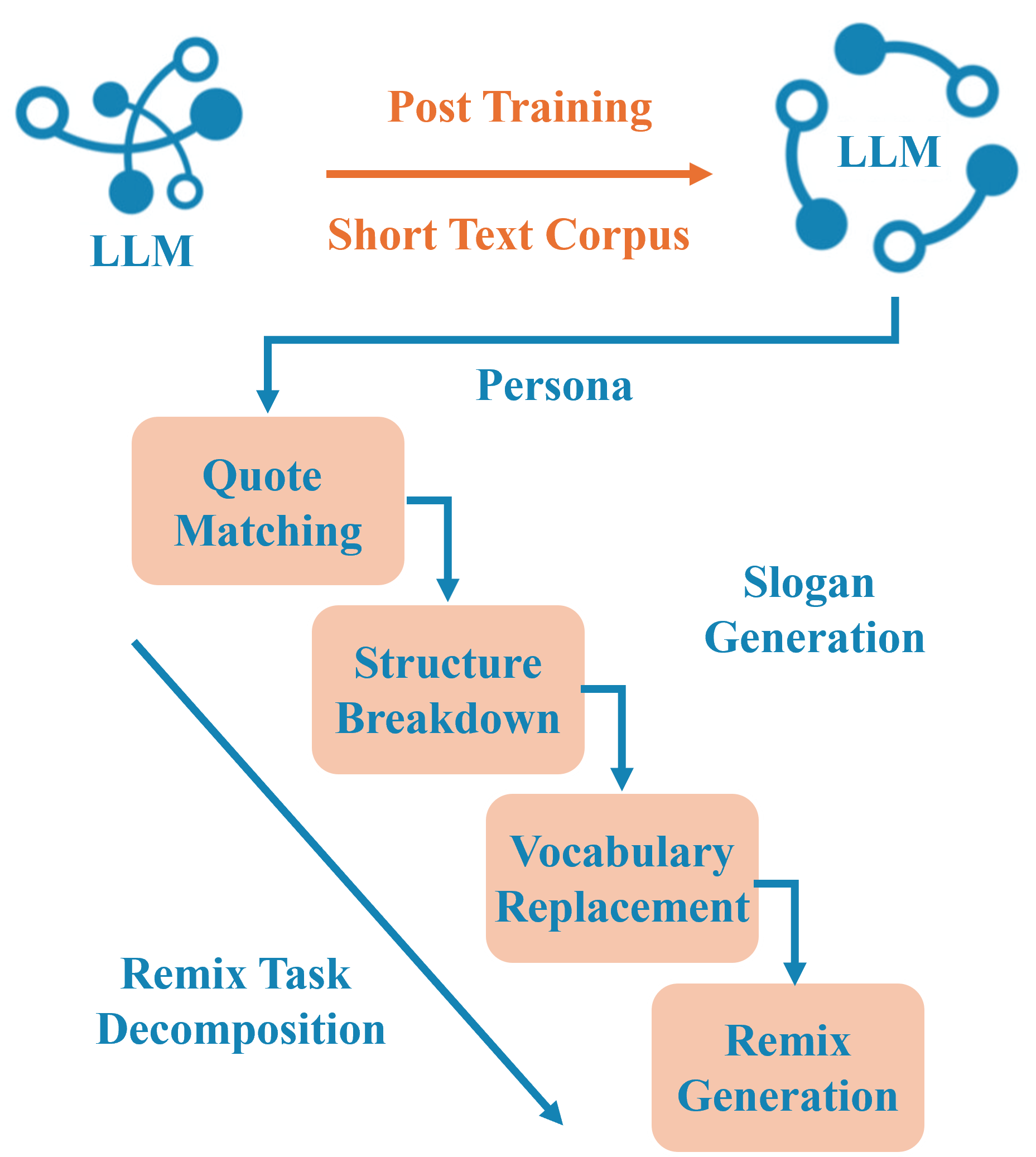}
  \caption{Framework of recontextualizing famous quotes for brand slogan generation.}\label{fig:framework}
\end{wrapfigure}
In this section, we present our framework for recontextualizing famous quotes to generate brand slogans. Given a brand or product and a target persona (e.g., proud, trustworthy, playful), our goal is to produce novel, memorable, and persona-aligned slogans. While we leverage the capabilities of LLMs, we do not rely on them to perform the entire task. Instead, we decompose the slogan remixing process into multiple subtasks, ensuring that each step is controllable, traceable, and produces high-quality outputs. Figure~\ref{fig:framework} illustrates the overall pipeline of our framework. The first subtask is continuous pre-training on a short-text corpus, designed to focus the LLM on concise text generation. The subsequent subtasks—quote matching, structure breakdown, vocabulary replacement, and remix generation—implement the concrete steps of the remix process. In the following, we provide a detailed explanation of each subtask.

\vspace{1mm}
\noindent\textbf{Post Training with Short Text Corpus}. Given an LLM, we first perform post-training on a carefully curated short-text corpus. Since slogans are inherently concise—typically under 20 words—our goal is to “reawaken” the LLM’s capacity for generating high-quality, memorable short text. To this end, we collect a diverse and culturally rich corpus that includes famous quotes, slang expressions, song lyrics, memes, and other popular short texts. This pre-training not only sharpens the model’s ability to produce coherent and engaging short phrases but also imbues it with stylistic and rhetorical diversity, laying a strong foundation for the subsequent subtask: aligning quotes with a target persona.

\noindent\textbf{Quote Matching}. This subtask focuses on identifying persona-aligned quotes that can serve as candidates for remixing. To ensure both quality and diversity, we prompt the LLM to return multiple candidate quotes—typically five or more—for each persona. By providing detailed persona specifications, the LLM can retrieve quotes that closely match the intended tone, style, and emotional resonance, ensuring that the selected candidates are well-suited for the subsequent remixing process. This step provides a strong foundation for slogan generation, particularly in aligning slogans with the target user group. For example, as illustrated in the introduction, fans of Percy Bysshe Shelley or Lu Xun are likely to resonate strongly with the remixed slogans derived from their works, enhancing both memorability and engagement.

\noindent\textbf{Structure Breakdown}. Based on the selected quotes from the previous subtask, we further identify which components must be preserved to maintain the original syntax and rhetorical structure, and which components can be modified during remixing. In other words, we use the LLM to decompose each quote into fixed segments and editable segments. The fixed segments retain the original sentence structure and rhetorical pattern, which are crucial to preserving recognizability and widely shared broader cultural resonance, while the editable segments offer flexibility to incorporate brand- or product-specific content.  

We continue to use the examples from the introduction for illustration. Percy Bysshe Shelley’s quote follows the structure “\textit{A comes, and B cannot be far behind},” while Lu Xun’s quote adopts the parallel structure “\textit{One is A, and the other is also A.}” Preserving these structural patterns is essential: if the underlying syntax is altered, the association with the original quote weakens, and the remix may fail to evoke the intended familiarity and insightfulness. This structured decomposition ensures that creativity is carefully and thoughtfully introduced without sacrificing recognizability, striking a balance between originality and memorability. 

\noindent\textbf{Vocabulary Replacement}. With the editable segments identified, we then generate candidate replacements while strictly preserving the original cadence and grammatical structure. In this subtask, we perform vocabulary replacement with minimal substitutions, ensuring that the remix remains closely aligned with the source quote. Specifically, we impose a hard constraint on the total number of changes from the original quote—for instance, inserting a brand or product name is counted as a modification. Additionally, we require the LLM to select replacement words or phrases with similar length and identical grammatical roles (e.g., noun-to-noun, verb-to-verb), which helps maintain rhythm, fluency, and stylistic consistency. This controlled replacement strategy carefully balances creativity with fidelity, enabling the generation of slogans that feel both novel and unmistakably grounded in their original quotes.

\noindent\textbf{Remix Generation}. Finally, we prompt the LLM to generate complete slogans in a remix style by integrating the preserved sentence structure with the substituted vocabulary. To further enhance quality and reliability, we include an additional refinement step in which the LLM polishes the generated slogans and verifies their logical coherence, semantic clarity, and content safety. This final validation stage serves as a safeguard against grammatical errors, semantic inconsistencies, or unintended or inappropriate messaging, ensuring that the resulting slogans are not only creative but also suitable for real-world deployment. During this process, some quotes selected in the initial stage may be discarded if the remixed outputs ultimately fail to satisfy these final quality checks.

\begin{table*}[t]
\centering
\caption{
\textbf{Illustrative examples of our generated slogans by remixing quotes.} Each remixed slogan adapts a well-known quote to a specific brand and persona while preserving strong first-glance emotional impact.
}\vspace{-2mm}
\resizebox{1.0\columnwidth}{!}{
\begin{tabular}{lc|cc}
\toprule
\textbf{Brand} & \textbf{Persona} & \textbf{Our Remixed Slogan} & \textbf{Original Quote} \\
\midrule

    Dawn & Joy & A splash of Dawn helps the grime go down! & A spoonful of sugar helps the medicine go down. \\
    
    Whirlpool & Joy & Whirlpool: Home is where the joy is. & Home is where the heart is. \\

    GE & Fear & The only thing we have to fear is missing GE. & The only thing we have to fear is fear itself. \\
    
   Clorox & Fear & Don’t wait until germs strike! & Don’t wait until it’s too late. \\
    
   DKNY & Anticipation & DKNY never sleeps. & New York never sleeps. \\
    
    Orgain & Anticipation & Something Organic is coming! & Something wonderful is coming. \\
    
    Apple & Pride & Stay bold, stay innovative. & Stay hungry, stay foolish. \\

    Philips & Pride & Trust is the foundation of every Philips solution. & Trust is the foundation of every great relationship. \\
    
    Aveeno & Trust & Trust is the glue of Aveeno. & Trust is the glue of life. \\

    Columbia & Trust & The safest Columbia gear leads to the highest destinations. & The safest path leads to the highest peaks. \\

\bottomrule
\end{tabular}
}
\label{tab:remix_examples}
\end{table*}

\begin{table*}[t]
\centering
\caption{
\textbf{Illustrative slogans generated by the state-of-the-art LLMs.} All slogans are generated using a standard brand–persona prompt that specifies the target brand and emotional tone, without additional recontextualizing. Due to the page limit, we only demonstrate the slogans generated by GPT-4o and DS-L. 
}\vspace{-2mm}
\resizebox{1.0\columnwidth}{!}{
\begin{tabular}{lc|cc}
\toprule
\textbf{Brand} & \textbf{Persona} 
& \textbf{Slogan Generated by GPT-4o } 
& \textbf{Slogan Generated by DS-L} \\
\midrule

Dawn & Joy 
& Awaken to Joy with Every Dawn!
& Dawn: Where Every Morning Shines.
 \\

Whirlpool & Joy 
& Whirlpool: Where Every Spin Sparks Joy!
& Whirlpool: Spinning Joy into Every Chore! 
 \\

GE & Fear 
& Don't Wait—Empower the Future with GE Today!
& GE Innovates Today for Tomorrow. 
 \\

Clorox & Fear 
& Don't Risk It—Stay Clean with Clorox!
& Clorox: Protect What You Can't See. 
 \\

DKNY & Anticipation 
& Discover Tomorrow with DKNY. 
& DKNY: What's Next? 
\\

Orgain & Anticipation 
& Fuel Your Tomorrow with Clean Nutrition.
& What's Next Starts Here. 
 \\

Apple & Pride 
& Innovation That Inspires Confidence.
& Pride in Every Innovation. 
 \\

Philips & Pride 
& Illuminate Your Excellence with Philips.
& Trusted by Generations for a Safer Tomorrow.
 \\

Aveeno & Trust 
& Aveeno: Trusted Care for Every Skin Journey.
& Aveeno: The Care You Can Trust.
 \\

Columbia  & Trust 
& Columbia: Where Reliability Meets Adventure.
& Columbia: Your Trusted Partner in Every Adventure.
 \\

\bottomrule
\end{tabular}
}
\vspace{-4mm}
\label{tab:llm}
\end{table*}

\section{Experiments}
In this section, we present our experimental results by first describing the experimental setup and then addressing the following research questions:
\begin{itemize}
\item \textit{What do slogans generated by remixing famous quotes look like?}
\item \textit{How do our remixed slogans perform compared with those generated by state-of-the-art LLM-based methods?}
\item \textit{Do our remixed slogans exhibit stronger attractiveness in human evaluations?}
\end{itemize}

\subsection{Experimental Setup}
We evaluate the effectiveness of our slogan generation framework under a controlled brand advertising setting. To ensure diversity across both brand and persona, we select a balanced set of 40 brands covering common marketing contexts in which slogan generation is typically applied. These brands span multiple consumer-facing domains, including beauty, baby, appliance, clothing, furniture, household, nutrition, and electronics. For each brand, we consider five representative personas: \textit{Pride}, \textit{Anticipation}, \textit{Fear}, \textit{Joy}, and \textit{Trust}.

\noindent\textbf{Baseline Methods}. We consider three LLMs for comparison: GPT-4o~\citep{hurst2024gpt}, DeepSeek-R1-Distill-LLaMA-70B, and DeepSeek-R1-Distill-Qwen-32B~\citep{guo2025deepseek}. For simplicity, we refer to the latter two models as DS-L and DS-Q, respectively.

\noindent\textbf{Implementation Details}. For our remix method, we first fine-tune QwQ-32B~\cite{qwen_qwq32b_blog} with parameter-efficient LoRA~\cite{hu2022lora} on a mixture of public text corpora covering slogans~\cite{slogan_dataset_kaggle,tenk_slogan_dataset_kaggle}, movie dialogs~\cite{danescu2011chameleons} and famous quotes~\cite{english_quotes_kaggle,quotes_500k_kaggle}. We combine and shuffle these datasets together and train for 3 epochs using AdamW~\cite{loshchilov2017decoupled} with a cosine learning-rate schedule with learning rate $1\times10^{-4}$, batch size 8, maximum sequence length 1,024, and bfloat16 precision. Then, the remix process is conducted on the above fine-tuned QwQ-32B model. 

For each \textit{brand--persona} combination, we generate a fixed set of $N$$=$$10$ candidate slogans from ours as well as the above baseline models. All models are prompted with the same brand and persona information to ensure a fair comparison.\footnote{All the prompts of our method for slogan generation and evaluation can be found in the Appendix.}


\noindent\textbf{Evaluation Metrics}. We conduct a comprehensive evaluation across multiple complementary dimensions, including \textit{diversity}, \textit{novelty}, \textit{emotional impact}, and \textit{human preference}. 

\textit{Diversity}. First, we assess slogan diversity using established metrics for open-ended text generation, including Distinct-2~\citep{li2016diversity}, Pairwise-BLEU, and Self-BLEU~\citep{papineni2002bleu, zhu2018texygen}. These metrics are widely adopted in prior work and jointly capture both local lexical variation and global distributional diversity across generated outputs~\citep{zhu2018texygen, tevet2021evaluating, zhang-etal-2025-evaluating-evaluation}.

Distinct-2 measures phrase-level lexical diversity by computing the proportion of unique word bi-grams among all generated slogans:
\begin{equation}\nonumber
    \mathrm{Distinct\text{-}2}
    =
    \frac{
    \left| \bigcup_{s \in \mathcal{S}} \mathrm{Ngram}_2(s) \right|
    }{
    \sum_{s \in \mathcal{S}} \left| \mathrm{Ngram}_2(s) \right|
    },
\label{eq:distinct2}
\end{equation}
where $s$ denotes a piece of generated slogan, and $\mathcal{S}$ presents the set of generated slogans of a method for the complete brand--persona combination and $\mathrm{Ngram}_2(s)$ is the set of bi-grams of $s$. A higher Distinct-2 value indicates richer phrase-level variation across the
generated slogans.

Pairwise-BLEU measures \textit{pairwise diversity} by quantifying the
average N-gram overlap between pairs of generated slogans. Higher overlap implies lower diversity; thus, pairwise BLEU serves as an inverse indicator of pairwise diversity. We compute the symmetrized sentence-level BLEU score for each unordered slogan pair and average all pairs:
\begin{equation}\nonumber
\mathrm{Pairwise\text{-}BLEU}
=
\frac{1}{\binom{N}{2}}
\sum_{1 \le i < j \le N}
\mathrm{BLEU}(s_i, s_j),
\label{eq:pairwise_bleu}
\end{equation}
where $\mathrm{BLEU}(s_i, s_j)$ denotes a symmetrized sentence-level BLEU score
between slogans $s_i$ and $s_j$.
Lower Pairwise BLEU values correspond to greater pairwise diversity among the slogans.

Self-BLEU captures diversity \textit{within a set of generated slogans} by
treating each slogan as a hypothesis and all remaining slogans as references,
and then averaging the resulting BLEU scores:
\begin{equation}\nonumber
\mathrm{Self\text{-}BLEU}
=
\frac{1}{N}
\sum_{i=1}^{N}
\mathrm{BLEU}
\left(
s_i,\,
\mathcal{S} \setminus \{s_i\}
\right).
\label{eq:self_bleu}
\end{equation}
Lower Self-BLEU indicates diverse generations with reduced overall repetition across outputs.

\begin{table*}[t]
    \centering
    \caption{
    \textbf{Diversity and novelty metrics for brand slogan generation.} Results are reported as mean $\pm$ standard deviation across all brand–persona combinations. An upward arrow ($\uparrow$) indicates that higher values are better, while a downward arrow ($\downarrow$) indicates that lower values are better. The best results are highlighted in bold.
    }
    \vspace{-2mm}
    \resizebox{0.8\columnwidth}{!}{
        \begin{tabular}{lccc|cc}
        \toprule

        \multirow{2}{*}{Model}& \multicolumn{3}{c}{\textbf{Diversity}} & \multicolumn{2}{c}{\textbf{Novelty}}\\
        \cline{2-6}

        & {Distinct-2} $\uparrow$ 
        & {Pairwise BLEU} $\downarrow$ 
        & {Self-BLEU} $\downarrow$ 
        & {Disfluency} $\downarrow$ 
        & {Unfaithfulness} $\downarrow$ \\
        \midrule
        
        GPT-4o 
        & 0.585 $\pm$ 0.135 
        & 0.098 $\pm$ 0.108 
        & 0.423 $\pm$ 0.200 
        & 4.05 $\pm$ 19.71
        & 33.45 $\pm$ 47.18 \\
        
        DS-L
        & 0.680 $\pm$ 0.123 
        & 0.060 $\pm$ 0.067 
        & 0.319 $\pm$ 0.184 
        & 13.45 $\pm$ 34.12 
        & 38.50 $\pm$ 48.66 \\
        
        DS-Q
        & 0.598 $\pm$ 0.138 
        & 0.104 $\pm$ 0.117 
        & 0.420 $\pm$ 0.210 
        & 8.55 $\pm$ 27.96 
        & 30.00 $\pm$ 45.83 \\

        \textbf{Ours} 
        & \textbf{0.840 $\pm$ 0.091} 
        & \textbf{0.013 $\pm$ 0.020} 
        & \textbf{0.099 $\pm$ 0.112} 
        & \textbf{36.95 $\pm$ 48.27} 
        & \textbf{57.55 $\pm$ 49.43} \\
        \bottomrule
        \end{tabular}
    }
    \vspace{-2.5mm}
    
    \label{tab:diversity_quality}
\end{table*}

\begin{figure*}[h]
    \centering
    \includegraphics[width=1.00\textwidth]{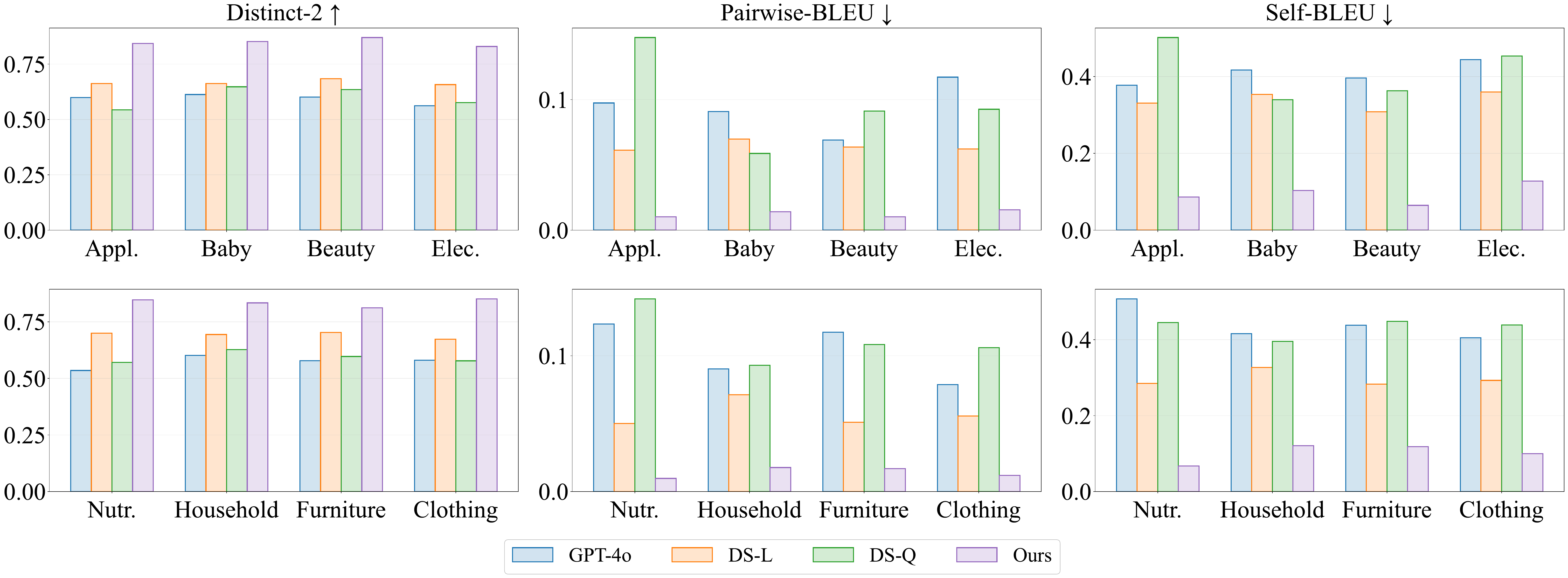}
    \vspace{-6mm}
    \caption{
    \textbf{Slogan diversity metrics across eight domains.} Each column corresponds to a diversity metric. The top row reports results for the appliance, baby, beauty, and electronics domains, while the bottom row reports results for the nutrition, household, furniture, and clothing domains.
    }
    \vspace{-4mm}
    \label{fig:diversity_bar}
\end{figure*}

In summary, given a set of generated slogans, these three diversity metrics capture complementary aspects of variation on different granularities. Distinct-2 measures local phrase-level diversity by quantifying how frequently word bi-grams are reused across slogans. Pairwise-BLEU assesses diversity at the pairwise level by evaluating the average similarity between slogan pairs. Self-BLEU captures set-level diversity by measuring how well each slogan can be explained by the remaining slogans in the set. Together, these metrics provide a holistic evaluation of slogan diversity.



\textit{Novelty}. Second, we evaluate the novelty of generated slogans using GPT-4o. Given the strong language modeling capabilities of modern LLMs, their outputs are typically grammatically correct and free of harmful content. However, highly fluent and faithful generations often closely align with the models’ internal knowledge distributions, which limits novelty. Following prior work~\cite{holtzman2019curious,chung2023increasing}, we assess novelty indirectly through measures of {Disfluency} and {Unfaithfulness}, where they are defined as 100 - Fluency and 100 - Faithfulness, respectively. Higher values indicate a greater departure from conventional or memorized expressions and, consequently, stronger novelty in slogan generation. 

\textit{Pairwise Emotional Impact}. Beyond these individual metrics, we perform pairwise comparisons using an \textit{LLM-as-a-Judge} framework to evaluate the first-glance emotional impact of the generated slogans by directly comparing the hook strength between two slogans~\citep{dolcos2017emerging, guitart2021impact, harini2025memorability}. 

\textit{Human Preference}. To ground our automatic evaluations in human perception, we also conduct a human preference study on the Prolific platform.

\subsection{Examples of Slogan Remix}
Table~\ref{tab:remix_examples} presents several examples of remixed slogans alongside their original quotes. The remixed slogans preserve the sentence structures of the original quotes, while also aligning closely with the target personas, demonstrating the effectiveness of our four-step remix pipeline. This approach not only reawakens the familiarity and resonance for audiences who know the original quotes but also introduces novelty and creativity through the remix.

For example, "\textit{A splash of Dawn helps the grime go down!}" and "\textit{Don’t wait until germs strike!}"—for Dawn and Clorox, two cleaning supply brands—effectively convey the Joy and Fear personas, respectively, while incorporating product-relevant words like "grime" and "germ." Similarly, the original quote "New York never sleeps," which evokes energy and youthfulness, is remixed for DKNY, a clothing brand, naturally inheriting that same lively feeling. The slogan "Something Organic is coming!" is remixed from "Something wonderful is coming," seamlessly linking the brand Orgain, a nutrition company, with a sense of wonder. Another notable example is "\textit{Stay bold, stay innovative}," originally inspired by "\textit{Stay hungry, stay foolish}," which carried a Fear persona. When adapting it for a Pride persona, our pipeline automatically substitutes more positive, aspirational words to reflect the desired persona while maintaining the structural and stylistic essence of the original quote. Remixing with quotes not only preserves insightfulness and memorability but also adds freshness and engagement.

We also present several slogans generated by baseline methods in Table~\ref{tab:llm} for comparison. While these slogans are generally coherent and make sense, they are similar to the existing slogans and tend to lack engagement and genuine insightfulness, which makes them less memorable.

\begin{figure*}[h]
    \centering
    \includegraphics[width=0.88\textwidth]{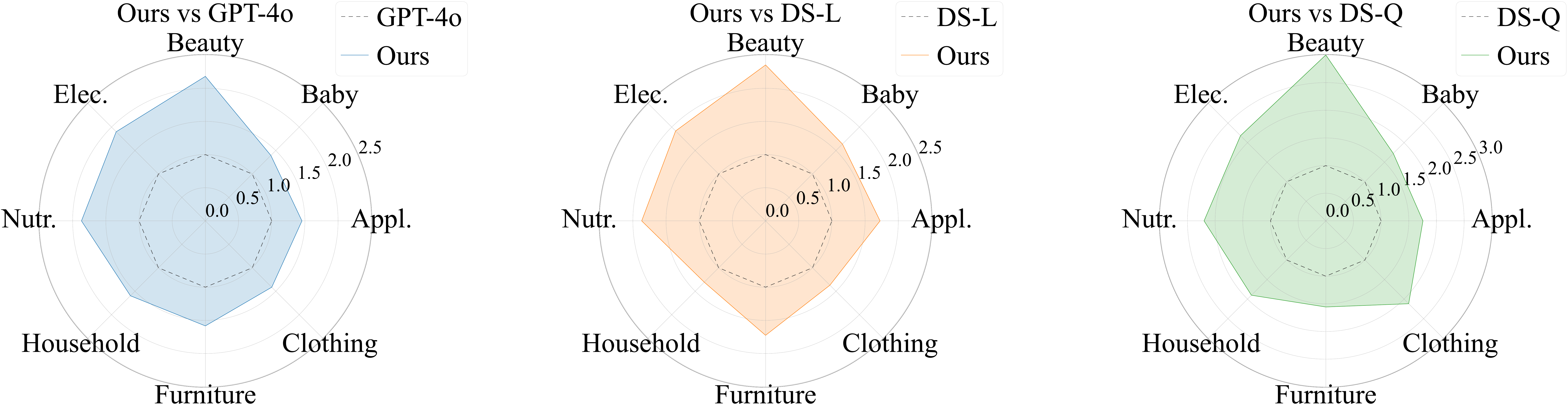}
    \caption{
    \textbf{Comparison of first-glance hook strength of our method against baselines.}
    Radar plots visualize the relative \textit{hook strength} of slogans generated by our method compared with three strong baseline models: GPT-4o, DS-L, and DS-Q.
    Each axis corresponds to one product domain, and values represent the normalized win ratio of our framework over the baseline, aggregated across all brand--persona combinations within that domain.
    }
    \vspace{-4mm}
    \label{fig:hook_radar}
\end{figure*}

\subsection{Performance by Quantitative Metrics}
Here we report the performance of our method and baselines in terms of diversity, novelty, and emotional impact across different settings. 

Table~\ref{tab:diversity_quality} reports the diversity and novelty metrics for brand slogan generation. Owing to the quote-remixing mechanism, our method achieves substantially higher diversity than all baseline approaches across multiple granularity levels. Notably, slogans generated by standard LLM baselines tend to be highly similar to one another, a trend clearly reflected in the diversity metrics. Specifically, GPT-4o and DS-Q achieve Distinct-2 scores of around 0.6, while DS-L remains below 0.7, whereas our remix-based method attains a Distinct-2 score above 0.8, significantly outperforming all baselines. Similar patterns are observed for Pairwise-BLEU and Self-BLEU, further confirming the superior diversity of our approach. Beyond the performance of all brand-persona combination reported in Table~\ref{tab:diversity_quality}, we also provide the diversity metrics across 8 domains in Figure~\ref{fig:diversity_bar}, which shares the similar phenomena that the slogans generated by our remixing method enjoy better diversity across multiple granularity levels and diverse domains. In terms of novelty, our method also outperforms LLM-generated slogans with an almost 20\% margin over the second best, exhibiting lower Disfluency and Unfaithfulness scores. This indicates that the generated slogans are less familiar to the LLM-based judge and deviate more from memorized or conventional expressions, thereby demonstrating stronger novelty.   

Conventional AI approaches, including large language models trained on extensive corpora of existing slogans, tend to generate outputs that closely resemble their training samples. While such methods can produce grammatically correct and semantically meaningful slogans, the resulting content often lacks originality and novelty, as it mirrors patterns already prevalent in the training data. This resemblance diminishes effectiveness over time due to advertising fatigue, where repeated exposure to similar slogans reduces user engagement and memorability. In contrast, our method deliberately avoids relying on existing slogan datasets. Instead, we broaden the creative scope by incorporating well-known quotes as the foundation for slogan generation. By recontextualizing these culturally familiar and rhetorically rich expressions, our approach injects additional creativity, insightfulness, and emotional resonance. This strategy not only mitigates advertising fatigue but also produces slogans that balance novelty with familiarity, making them more engaging, memorable, and impactful in real-world branding scenarios.

Beyond linguistic diversity and novelty, effective brand slogans must also demonstrate a strong \textit{first-glance emotional impact}, also known as a hook that captures attention immediately. Such hook strength is particularly important in real-world advertising scenarios, where audiences often engage with brand messages fleetingly and make rapid, intuition-driven judgments rather than careful semantic parsing. Prior work in advertising and cognitive psychology has shown that emotionally evocative content is more likely to capture attention, enhance memorability, and influence downstream engagement and decision-making~\citep{dolcos2017emerging, guitart2021impact, harini2025memorability}. Following these literatures, we explicitly evaluate \textit{hook strength}, defined as the immediate emotional hook of a slogan at first glance. We design a pairwise \textit{LLM-as-a-Judge} protocol with GPT-4o that instructs the judge to react as a human viewer, prioritizing surprise, humor, attitude, and controlled boldness. Given a brand, a target persona, and two slogans, the judge selects the slogan that delivers a stronger immediate emotional punch, or declares a tie if both feel equally impactful. 

Let $W$, $L$, and $T$ denote the numbers of wins, losses, and ties of one method over another one, then the hook score can be calculated as $(W+T/2)/(L+T/2)$. Figure~\ref{fig:hook_radar} visualizes the pairwise hook score ratio of our recontextualizing framework against three strong baseline models under the hook-based pairwise evaluation protocol. Each axis corresponds to a product domain, and the dashed line represents the self-comparison baseline, where all values are equal to 1. Our framework consistently outperforms or remains competitive with all baselines, including GPT-4o, the judge,  across most product categories, with particularly pronounced gains in beauty, nutrition, and electronic domains. These results indicate that our recontextualizing slogan-generation pipeline through remix and controlled deviation encourages brand slogans that are more emotionally salient and attention-grabbing. 

\subsection{Human Evaluation}
To evaluate our methods with real users, we run a small-scale crowd study on Prolific. We start from 40 brands and 5 personas (200 brand--persona scenarios) and generate slogans for all methods on all scenarios. We then shuffle all 200 scenarios with a fixed random seed, and select 20 questions to form the questionnaire. This gives a diverse but still manageable subset that covers multiple brands and personas while keeping the per-participant workload reasonable. Each question specifies a brand and a persona, and shows four candidate slogans (one from each method; the order of options is randomized). For every question, participants make two single-choice judgments:
(1) \textit{Hits you first}, choosing the slogan that grabs them most at first glance; and
(2) \textit{Persona-fit best}, choosing the slogan that best matches the given persona.


In this run, we recruit 10 participants, giving 200 total judgments for each question type (10$\times$20). We include an attention-check question with an unambiguous correct answer and would exclude any participants who fail it; in this run, the attention-check failure rate is 0\%. We aggregate human choices by counting how often each method is selected. On \textit{Hits you first}, our method receives 59 out of 200 selections (29.5\%), compared to 53/200 (26.5\%) for DS-Q, 49/200 (24.5\%) for DS-L, and 39/200 (19.5\%) for GPT-4o. On \textit{Persona-fit best}, DS-Q is selected most often (61/200, 30.5\%), followed by DS-L (48/200, 24.0\%) and GPT-4o (48/200, 24.0\%), while our method receives 43/200 (21.5\%). Overall, these results suggest that our remix-based method is competitive at grabbing attention at first glance, which is also consistent with our above quantitative evaluation, while DS-Q remains slightly better aligned with the target personas in terms of human preference.

\section{Conclusion}
In this work, we propose a novel paradigm for brand slogan generation by recontextualizing persona-related famous quotes. Unlike conventional data-driven approaches that learn directly from existing slogans and are therefore prone to generating repetitive and less engaging content, our framework expands the creative scope to a broader quote domain, effectively mitigating advertisement fatigue. By decomposing the remix process into controllable and traceable subtasks—including quote matching, structural preservation, vocabulary replacement, and remix generation—we leverage the strengths of large language models while maintaining fine-grained control over creativity and quality. Extensive experiments across automatic metrics and human evaluations demonstrate that our method consistently produces slogans with higher diversity, stronger novelty, and greater emotional impact than state-of-the-art LLM-based baselines. The remixed slogans not only preserve the rhetorical structure and insightfulness of well-known quotes but also align closely with brand personas, making them both memorable and engaging. Overall, our work highlights the importance of moving beyond slogan-centric training data and provides a practical, interpretable, and scalable framework for creative slogan generation in real-world advertising systems.

\clearpage
\bibliographystyle{unsrtnat}

\bibliography{references}

\clearpage
\appendix
\onecolumn

\newpage
\appendix
\onecolumn
\section*{Appendix}
In the appendix, we provide all the prompts of our method for slogan generation and evaluation.

\subsection*{Prompts for our framework for slogan remix}
Table 4 shows the prompt for our remix method to generate slogans. 

\begin{table}[h]
    \centering
    \caption{\small
    Prompt card for the \textit{slogan remix} generator used in our framework. Steps, output template, and constraints mirror our implementation.
    }
    \label{tab:prompt_remix}
    \small
    \begin{tabular}{|>{\raggedright\arraybackslash}p{7.25cm}|}
    \midrule
    \textcolor{prompt1}{Role Definition:}
    You are an award-winning slogan copywriter.
    Your task is to remix a classic quote, slang, lyric, meme, or popular expression into a brand slogan.
    Follow ALL steps and output in English. \\

    \midrule
    \textcolor{prompt2}{Input Context:} \\
    \textbf{Brand:} \textit{\{Brand\}} \\
    \textbf{Emotion (tone only):} \textit{\{Persona\}} \\
    \textbf{Emotion goal:} \textit{\{persona\_guidelines[Persona]\}} \\

    \midrule
    \textcolor{prompt3}{Procedure (Steps):} \\
    \begin{minipage}[t]{\linewidth}
    \vspace{-2mm}
    \begin{itemize}[leftmargin=*, itemsep=0pt, topsep=0pt, parsep=0pt, partopsep=0pt]
        \item \textbf{Step 1 (Quote Matching).}
        Propose three classic sentences (quote/slang/lyric/meme), each 5--10 English words, each with author; 
        do not state the emotion name; briefly explain why each fits the emotion and brand; 
        mark the best with star and repeat it on a new line with an explanation for selection.
        \item \textbf{Step 2 (Structure Breakdown).}
        Copy the star quote and use \texttt{|} to separate \emph{editable} vs \emph{fixed} parts; 
        indicate which words can be replaced by brand keywords and why.
        \item \textbf{Step 3 (Vocabulary Replacement).}
        Propose 1--2 replacement word/phrase sets with short reasons; 
        change at most two words in total (adding the brand name counts toward this limit).
        \item \textbf{Step 4 (Remix Generation).}
        Using the best replacement, output \emph{ONE} final slogan that keeps original syntax/meter, 
        includes the brand name/keyword, and ends with punctuation; 
        after the slogan, output \texttt{END\_OF\_REMIX} on a new line.
    \end{itemize}
    \end{minipage} \\

    \midrule
    \textcolor{prompt4}{Output Format (verbatim):} \\
    \texttt{Step 1 Quote Matching: ...}\\
    \texttt{Step 2 Structure Breakdown: ...}\\
    \texttt{Step 3 Vocabulary Replacement: ...}\\
    \texttt{Step 4 Remix Slogan: "..."}\\
    \texttt{END\_OF\_REMIX} \\

    \midrule
    \textcolor{prompt5}{Important Constraints:} \\
    Do \textbf{not} use first-person words (I, we, me, my, our). 
    Use neutral or second-/third-person perspective so the brand is mentioned, not the narrator. \\

    \midrule
    \end{tabular}
    \vspace{-2mm}
    
    \vspace{-5mm}
\end{table}

\subsection*{Prompts for slogan evaluation}
Table 5 \& 6 show the prompt for faithfulness evaluation and pairwise hook-based evaluation using an LLM-as-a-Judge framework. The prompt for fluency is similar to the one of faithfulness, which we omit it here.

\begin{table}[h]
    \centering
    \caption{\small
    An example prompt card for \textit{faithfulness evaluation} using an LLM-as-a-Judge framework.
    The prompt checks whether the (brand, persona) question context implies the given slogan.
    }
    \label{tab:prompt_faithfulness}
    \small
    \begin{tabular}{|>{\raggedright\arraybackslash}p{7.25cm}|}
    \midrule
    \textcolor{prompt1}{Role Definition:}
    You are a careful evaluator. Follow the instructions exactly and do not add extra text.
    Your output must contain exactly two lines. \\

    \midrule
    \textcolor{prompt2}{Evaluation Guidelines:} \\
    \begin{minipage}[t]{\linewidth}
    \vspace{-2mm}
    \begin{itemize}[leftmargin=*, itemsep=0pt, topsep=0pt, parsep=0pt, partopsep=0pt]
        \item Judge whether the \textbf{question text implies the ad text}.
        \item Answer \textbf{1} if it implies; answer \textbf{0} if it does not.
        \item Focus on semantic consistency; do not add assumptions not present in the question text.
    \end{itemize}
    \end{minipage} \\

    \midrule
    \textcolor{prompt3}{Input Context:} \\
    \textbf{Question text:} \\
    Brand: \textit{\{Brand\}} \\
    Persona: \textit{\{Persona\}} \\[0.5mm]
    \textbf{Ad text:} \\
    \textit{\{Slogan\}} \\

    \midrule
    \textcolor{prompt4}{Task:} \\
    Decide whether the question text implies the ad text. \\

    \midrule
    \textcolor{prompt5}{Output Format:} \\
    Please answer in exactly two lines: \\
    \textbf{Answer:} 0 / 1 \\
    \textbf{Reason:} One short sentence. \\

    \midrule
    \end{tabular}
    \vspace{-2mm}
    
    \vspace{-5mm}
\end{table}

\begin{table}[h]
    \centering
    \caption{\small
    An example prompt card for \textit{pairwise hook-based evaluation} using an LLM-as-a-Judge framework.
    The prompt is explicitly designed to simulate a human's first-glance emotional reactions to brand slogans.
    }
    \label{tab:prompt_hook}
    \small
    \begin{tabular}{|>{\raggedright\arraybackslash}p{7.25cm}|}
    \midrule
    \textcolor{prompt1}{Role Definition:} 
    You are a human reacting instantly to brand slogans. 
    You do \textbf{not} analyze professionally and do \textbf{not} consider brand safety, approval processes, or marketing rules.
    Your judgment is based solely on immediate emotional impact and gut reaction at first glance. \\
    
    \midrule
    \textcolor{prompt2}{Evaluation Guidelines:} \\
    \begin{minipage}[t]{\linewidth}
    \vspace{-2mm}
    \begin{itemize}[leftmargin=*, itemsep=0pt, topsep=0pt, parsep=0pt, partopsep=0pt]
        \item Evaluate solely based on \textbf{first-glance emotional punch}.
        \item Prefer slogans that are surprising, humorous, or edgy.
        \item Discourage generic, conservative, or overly polished slogans.
        \item Controlled boldness and mild risk-taking are accepted.
    \end{itemize}
    \end{minipage} \\
    \midrule
    \textcolor{prompt3}{Input Context:} \\
    \textbf{Brand:} \textit{\{Brand\}} \\
    \textbf{Target Persona:} \textit{\{Persona\}} \\[0.5mm]
    \textbf{Slogan A:} \\
    \textit{\{Slogan A\}} \\[0.5mm]
    \textbf{Slogan B:} \\
    \textit{\{Slogan B\}} \\

    \midrule
    \textcolor{prompt4}{Task:} \\
    Decide which slogan triggers a stronger \textbf{immediate emotional reaction}.
    If both slogans feel equally strong or equally weak, answer C. \\

    \midrule
    \textcolor{prompt5}{Output Format:} \\
    Please answer in exactly two lines: \\
    \textbf{Answer:} A / B / C \\
    \textbf{Reason:} One short phrase describing your gut reaction. \\

    \midrule
    \end{tabular}
    \vspace{-2mm}
    
    \vspace{-5mm}
\end{table}

\end{document}